\documentclass{article}

\usepackage{microtype}
\usepackage{graphicx}
\usepackage{subfigure}
\usepackage{booktabs} 
\usepackage[font=small]{caption}
\usepackage{multirow,array,amsmath,amssymb,xcolor}
\usepackage[acronym]{glossaries}

\usepackage{hyperref}

\usepackage[accepted]{mlsys2025_arxiv} 

\usepackage{enumitem}

\setlist{
    nosep,
    topsep=2pt,
    partopsep=0pt,
    itemsep=1pt,
    parsep=0pt
}

\newcommand{\Ds}{\mathcal{S}}\newcommand{\Do}{\mathcal{O}}\newcommand{\Da}{\mathcal{A}}
\newcommand{\status}[1]{\textit{\small[#1]}}
\newcolumntype{L}[1]{>{\raggedright\arraybackslash}p{#1}}
\mlsystitlerunning{A Capability Framework for Evaluating Agentic AI Platforms in Regulated Enterprises}

\makeglossaries

\begin{document}

\newacronym{Al-Jazeera}{Al-Jazeera}{Al-Jazeera Satellite Channel}
\newacronym{AraGPT-2}{AraGPT-2}{Arabic Generative Pre-trained Transformer 2}
\newacronym{AraGPT-3}{AraGPT-3}{Arabic Generative Pre-trained Transformer 3}
\newacronym{ArSAS}{ArSAS}{Arabic Speech-Act and Sentiment}
\newacronym{PATB}{PATB}{Penn Arabic Treebank}
\newacronym{BBC}{BBC}{British Broadcasting Corporation}
\newacronym{BERT}{BERT}{Bidirectional Encoder Representations from Transformers}
\newacronym{BPE}{BPE}{byte-pair encoding}
\newacronym{mBERT}{mBERT}{multi-lingual BERT}
\newacronym{RoBERTa}{RoBERTa}{A Robustly Optimized BERT Pre-training Approach}
\newacronym{CA}{CA}{Classical Arabic}
\newacronym{CC}{CC}{Common Crawl}
\newacronym{CC-100}{CC-100}{Common Crawl 100}
\newacronym{CNN}{CNN}{Cable News Network}
\newacronym{DA}{DA}{Dialectal Arabic}
\newacronym{EAPCOUNT}{EAPCOUNT}{English-Arabic Parallel Corpus of the United Nations Texts}
\newacronym{GPT-1}{GPT-1}{Generative Pre-trained Transformer 1}
\newacronym{GPT-2}{GPT-2}{Generative Pre-trained Transformer 2}
\newacronym{GPT-3}{GPT-3}{Generative Pre-trained Transformer 3}
\newacronym{GPT}{GPT}{Generative Pre-trained Transformer}
\newacronym{GPU}{GPU}{Graphics Processing Unit}
\newacronym{HPL}{HPL}{High-Performance Linpack} 
\newacronym{IR}{IR}{Information Retrieval} 
\newacronym{KSUCCA}{KSUCCA}{King Saud University Corpus of Classical Arabic}
\newacronym{LM}{LM}{Language model}
\newacronym{LLM}{LLM}{Large Language Model}
\newacronym{LSTM}{LSTM}{Long-Short Term Memory}
\newacronym{ME}{ME}{Middle East}
\newacronym{MGB-2}{MGB-2}{Multi-Genre Broadcast-2}
\newacronym{MSA}{MSA}{Modern Standard Arabic}
\newacronym{MEGLN}{MEGLN}{MSA, Egyptian, Gulf, Levant \& North African}
\newacronym{MRC}{MRC}{Machine Reading Comprehension}
\newacronym{MT}{MT}{Machine Translation}
\newacronym{NN}{NN}{Neural Network}
\newacronym{NQ}{NQ}{Natural Questions}
\newacronym{NLP}{NLP}{Natural Language Processing}
\newacronym{OpenITI}{OpenITI}{Open Islamicate Texts Initiative}
\newacronym{OSAC}{OSAC}{Open-Source Arabic Corpora}
\newacronym{OSCAR}{OSCAR}{Open Super-large Crawled ALMAnaCH coRpus}
\newacronym{OSIAN}{OSIAN}{Open Source International Arabic News}
\newacronym{POS}{POS}{Part-of-speech}
\newacronym{PTB}{PTB}{Penn Tree Bank}
\newacronym{QA}{QA}{Question Answering} 
\newacronym{RNN}{RNN}{Recurrent Neural Network}
\newacronym{RT}{RT}{Russia Today}
\newacronym{SM}{SM}{Sentence Match}
\newacronym{SQuAD}{SQuAD}{Stanford Question Answering Dataset}
\newacronym{TPU}{TPU}{Tensor Processing Unit}
\newacronym{TFRC}{TFRC}{TensorFlow Research Cloud}
\newacronym{KSU}{KSU}{King Saud University}
\newacronym{WER}{WER}{Word Error Rate}
\newacronym{XNLI}{XNLI}{Cross-lingual Natural Language Inference}
\newacronym{HARD}{HARD}{Hotel Arabic Reviews Dataset}

\twocolumn[
\mlsystitle{A Capability Framework for Evaluating Agentic AI Platforms in Regulated Enterprises}

\mlsyssetsymbol{equal}{*}

\begin{mlsysauthorlist}
\mlsysauthor{Abbas Raza Ali}{equal,abbas}
\mlsysauthor{Muhammad Ajmal Siddiqui}{ajmal}
\mlsysauthor{Moona Zahid}{moona}
\end{mlsysauthorlist}

\mlsysaffiliation{abbas}{Citigrounp Inc., London, United Kingdom}
\mlsysaffiliation{ajmal}{Ernst \& Young LLP, London, United Kingdom}
\mlsysaffiliation{moona}{NVIDIA Corporation, London, United Kingdom}

\mlsyscorrespondingauthor{Abbas Raza Ali}{abbas.raza.ali@gmail.com}

\mlsyskeywords{Agentic AI, Multi-agent systems, Enterprise evaluation, Governance, Benchmarking}

\vskip 0.3in

\begin{abstract}
Agentic AI platforms-systems in which LLM agents plan, invoke tools, retain memory and coordinate to change the state of enterprise systems-are being procured faster than they can be measured. Public agent benchmarks score \emph{models} in fixed harnesses; enterprises buy integrated \emph{platforms} whose orchestration, memory, governance and tooling layers dominate outcomes, and they buy them largely on the strength of vendor demonstrations. We report an industrial evaluation programme (``bakeoff'') run by the innovation laboratory of a global bank in which ten platforms-spanning open-source frameworks, proprietary products and hyperscaler ecosystems-were assessed by a cross-functional panel on nine capability dimensions across three financial-services workloads: counterparty credit risk, retrieval-augmented generation and process automation. Scores separate the field into three tiers that are stable under alternative weightings, and surface workload-specific strengths that aggregates hide. A secondary analysis, however, finds Cronbach's $\alpha=0.98$ across nominally distinct dimensions and rank correlations $\geq 0.93$ with the total for eight of nine: a halo effect indicating that demonstration-based scoring measures presentation as much as capability. From these lessons we propose a \emph{gate-then-score} evaluation protocol with four layers (capability, execution reliability, governance gates, business value), trace-level reliability metrics estimated over repeated runs, an autonomy-readiness scale and a reproducibility protocol. We state precisely what the present evidence supports, and what a controlled harness must add to turn enterprise platform evaluation from structured judgement into measurement.
\end{abstract}
]

\printAffiliationsAndNotice{}

\section{Introduction}
Within three years, LLM agents have progressed from prompting patterns that interleave reasoning and action \citep{yao2023react,wei2022cot} to systems that resolve real software issues \citep{jimenez2024swebench}, operate desktop environments \citep{xie2024osworld} and coordinate as teams of specialised roles \citep{wu2023autogen,hong2024metagpt,fourney2024magentic}. Vendors have responded with a rapidly expanding catalogue of ``agent platforms'', and interoperability protocols-the Model Context Protocol (MCP) \citep{anthropic2024mcp} and Agent2Agent (A2A) \citep{google2025a2a}-promise a common substrate for connecting agents to tools and to each other. For a global bank the question is no longer whether such systems are interesting but which of them can be trusted to formulate and execute sequences of \emph{state-changing} actions inside a regulated operating environment.

That question is hard for three reasons. First, the benchmark literature measures models, not platforms: AgentBench \citep{liu2024agentbench}, GAIA \citep{mialon2024gaia}, WebArena \citep{zhou2024webarena} and $\tau$-bench \citep{yao2024taubench} score an LLM's behaviour in a fixed harness, whereas an enterprise buys a product whose orchestration, memory, governance and tooling layers it does not control. Second, vendor claims outrun evidence: \citet{kapoor2024matter} document that many agent evaluations are irreproducible, ignore cost and overfit leaderboards. Third, regulated institutions must satisfy model-risk, auditability and accountability obligations \citep{fed2011sr117,nist2023rmf,euaiact2024} that no public benchmark addresses. Buyers therefore fall back on vendor demonstrations-with the attendant risk that presentation quality is mistaken for engineering quality.

This paper reports on a programme designed to confront that gap: an ``Agentic AI Bakeoff'' run by the innovation laboratory of a global financial institution (hereafter ``the Lab''). We treat it both as an industrial case study and as an opportunity to formalise a reusable evaluation methodology, and we are explicit about which parts of the paper are \status{established}, \status{documented}, \status{observed}, \status{proposed} or \status{hypothesis}.

\paragraph{Research questions.}
\textbf{RQ1} Which system-level capabilities, beyond model quality, determine whether an agentic platform is deployable in a regulated enterprise?
\textbf{RQ2} How can they be measured in a manner that is scenario-based, traceable and reproducible rather than demonstration-based?
\textbf{RQ3} What patterns emerge when a multi-vendor bakeoff is scored against such a rubric across risk-analysis, retrieval and automation workloads?
\textbf{RQ4} What additional evidence is required to move from an industrial evaluation to a controlled benchmark study?

\paragraph{Contributions.}
(i)~A conceptual grounding that positions agentic AI relative to RPA and single agents, with a working definition, a sequential-decision formalism and a reference architecture mapped to the literature (\S\ref{sec:background}).
(ii)~A nine-dimension capability taxonomy, a multi-stage protocol and a scoring model with reliability diagnostics (\S\ref{sec:framework}).
(iii)~Empirical results from ten platforms on three financial-services scenarios, with a psychometric analysis of the scores and an explicit threats-to-validity assessment (\S\ref{sec:results}--\ref{sec:threats}).
(iv)~A \emph{gate-then-score} layered evaluation model with trace-level metrics and a pseudocode protocol (\S\ref{sec:layers}), an autonomy-readiness scale and a business-value identity (\S\ref{sec:discussion}), and a research agenda specifying the controlled experiments still required. Platform profiles, an agentic document-extraction case study, scoring anchors and a reproducibility protocol are in the supplementary appendix (separate PDF).

\section{Background and Formalisation}\label{sec:background}
\subsection{The automation continuum}
\begin{figure}[t]\centering\includegraphics[width=\columnwidth]{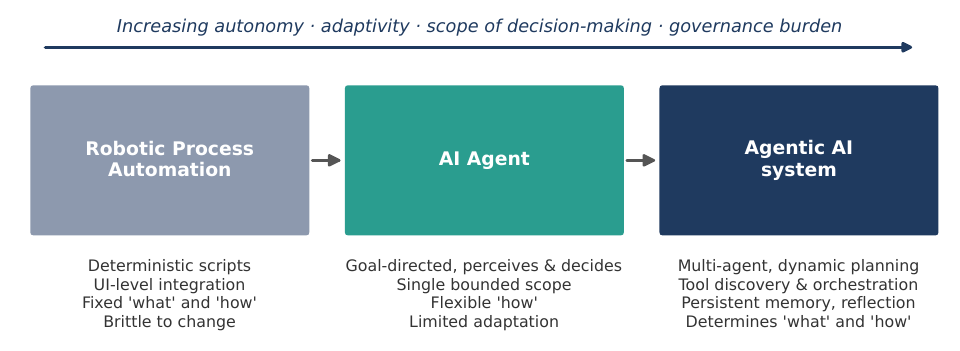}
\caption{The automation continuum. Rightward movement increases run-time autonomy, the scope of delegated decisions and, consequently, the governance burden.}\label{fig:continuum}\end{figure}

\emph{RPA}, \emph{AI agent} and \emph{agentic AI} are routinely conflated. They occupy distinct positions on a continuum defined by how much of the ``what'' and the ``how'' of a task is decided at run time rather than fixed at design time (Fig.~\ref{fig:continuum}). RPA bots mimic human interaction with application UIs to execute rule-based work \citep{aalst2018}; they are deterministic and non-invasive but brittle to UI change. A classical agent perceives its environment and acts toward goals \citep{wooldridge1995,russell2020} within one bounded task. We use \emph{agentic AI} for systems-often multi-agent-that dynamically plan, adapt and collaborate toward open-ended goals, discovering and invoking tools and maintaining memory across interactions \citep{wang2024survey,xi2023rise}. A further practical distinction: an \emph{agentic workflow} fixes control flow at design time and lets the model fill adaptive steps, whereas an \emph{agentic system} lets the model direct control flow \citep{anthropic2024agents}. Most enterprise deployments we observed were workflows with agentic steps; the evaluation problem is partly a design-selection problem in which the target is the \emph{minimum} autonomy that reliably achieves the outcome.

\subsection{Working definition}
The Lab's parent institution defines agentic AI as systems that \emph{achieve goals on behalf of a user or another system by autonomously formulating and executing a sequence of actions that invoke other systems and change their state}. Two clauses carry the weight: (a)~autonomous formulation and execution of an action sequence, and (b)~state-changing invocation of other systems. Clause (b) separates agentic AI from conversational assistants and raises its risk profile: errors propagate into records, workflows and downstream decisions \citep{chan2023harms,shavit2023governing}.

\subsection{A formal characterisation}\label{sec:formal}
In the vocabulary of sequential decision-making \citep{kaelbling1998,russell2020}, an agentic system operates in an environment with latent state $s\in\Ds$, receives observations $o\in\Do$ and selects actions $a\in\Da=\Da_{\text{tool}}\cup\Da_{\text{msg}}\cup\Da_{\text{text}}$ (tool invocations, messages to agents or humans, text outputs). A single agent is a policy with explicit memory state $m_t$:
\begin{equation}
a_t=\pi(o_t,m_t;\theta),\qquad m_{t+1}=\mu(m_t,o_t,a_t),\label{eq:policy}
\end{equation}
where $\mu$ is a memory-management function (in LLM agents, context management plus external stores \citep{packer2023memgpt,sumers2024coala}) and $\theta$ the underlying model. Tool actions induce transitions $s_{t+1}\sim T(\cdot\mid s_t,a_t)$, formalising clause~(b). Planning selects $a_1\ldots a_n$ to maximise expected goal attainment; reflection is an inner loop in which the agent evaluates its trajectory and revises behaviour through verbal feedback rather than gradient updates \citep{shinn2023reflexion}. A multi-agent system is a set $\{\pi^i\}$ with communication topology $\Gamma$ and an orchestration function routing sub-goals; most multi-agent failures arise in specification and coordination rather than in individual reasoning \citep{cemri2025fail}. The formalism is standard; its role here is to state which component each evaluation dimension probes.

\subsection{Reference architecture}
\begin{figure}[t]\centering\includegraphics[width=\columnwidth]{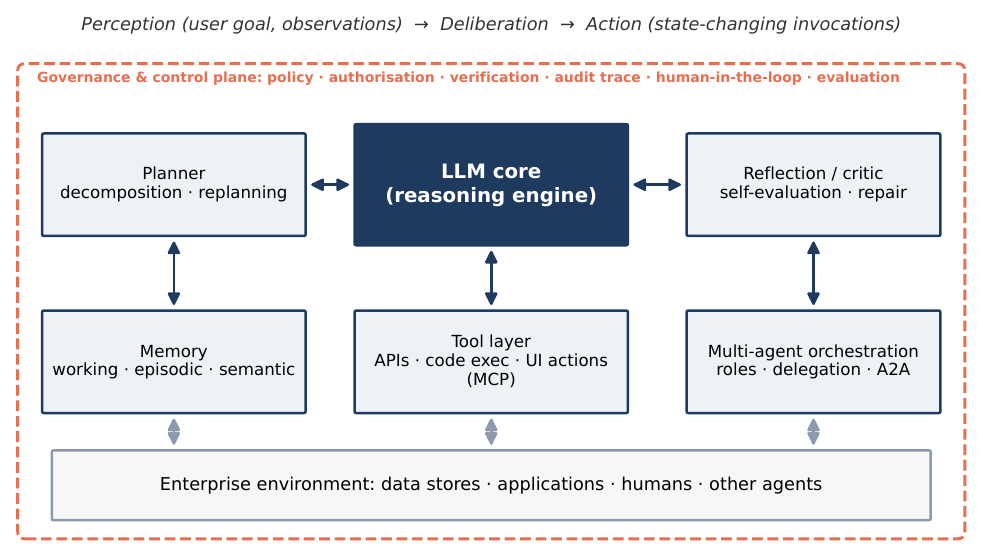}
\caption{Reference architecture for an enterprise agentic platform. The dashed boundary is the governance and control plane that regulated deployments must wrap around every component.}\label{fig:arch}\end{figure}

Figure~\ref{fig:arch} abstracts the platforms we evaluated, consistent with the cognitive-architecture framing of \citet{sumers2024coala} and recent pattern surveys \citep{masterman2024landscape,anthropic2024agents}. Each component maps to an active research thread: reasoning \citep{wei2022cot,yao2023tot,yao2023react}; planning \citep{wang2024voyager}; tool use \citep{schick2023toolformer,patil2023gorilla,qin2024toolllm,yan2024bfcl}; memory, motivated by finite and unevenly used context \citep{liu2024lost,packer2023memgpt,park2023generative}; reflection, whose unaided form is unreliable without external feedback \citep{shinn2023reflexion,huang2024selfcorrect}; multi-agent orchestration \citep{wu2023autogen,hong2024metagpt,li2023camel,fourney2024magentic}; and a governance plane demanded by risk frameworks \citep{nist2023rmf,nist2024genai,euaiact2024} and by the literature on agentic harms, including indirect prompt injection through tool outputs \citep{greshake2023injection,debenedetti2024agentdojo}.

\section{Related Work}\label{sec:related}
Agent evaluation has developed along two tracks. \emph{Model-level} benchmarks-AgentBench, GAIA, WebArena, SWE-bench, OSWorld-fix a harness and vary the model. \emph{Enterprise-workload} benchmarks are more recent: WorkArena evaluates knowledge-work tasks on a live ServiceNow instance \citep{drouin2024workarena}; $\tau$-bench adds multi-turn user interaction under policy constraints and the pass$^k$ reliability metric \citep{yao2024taubench}; AgentArch compares agent \emph{architectures} on enterprise tasks and reports at most $\approx$35\% success on its harder task \citep{agentarch2025}; EnterpriseOps-Gym tests multi-step operations with irreversible actions \citep{enterpriseops2026}; DRBench targets enterprise deep research \citep{abaskohi2026drbench}; and a multi-dimensional evaluation framework for enterprise agents has been proposed \citep{beyondaccuracy2025}. AgentDojo adds adversarial robustness, including banking tasks \citep{debenedetti2024agentdojo}. Measurement-quality critiques \citep{kapoor2024matter,stroebl2024flaws} and the long-horizon analysis of \citet{kwa2025metr} motivate our reliability layer. This paper is complementary: it is a \emph{buyer-side, platform-level} evaluation of commercial products as delivered, inside a regulated institution, with the organisational and governance attributes that public benchmarks abstract away. Its contribution is the protocol and its diagnostics rather than a new task suite.

\section{Evaluation Framework}\label{sec:framework}
\begin{figure*}[t]\centering\includegraphics[width=0.92\textwidth]{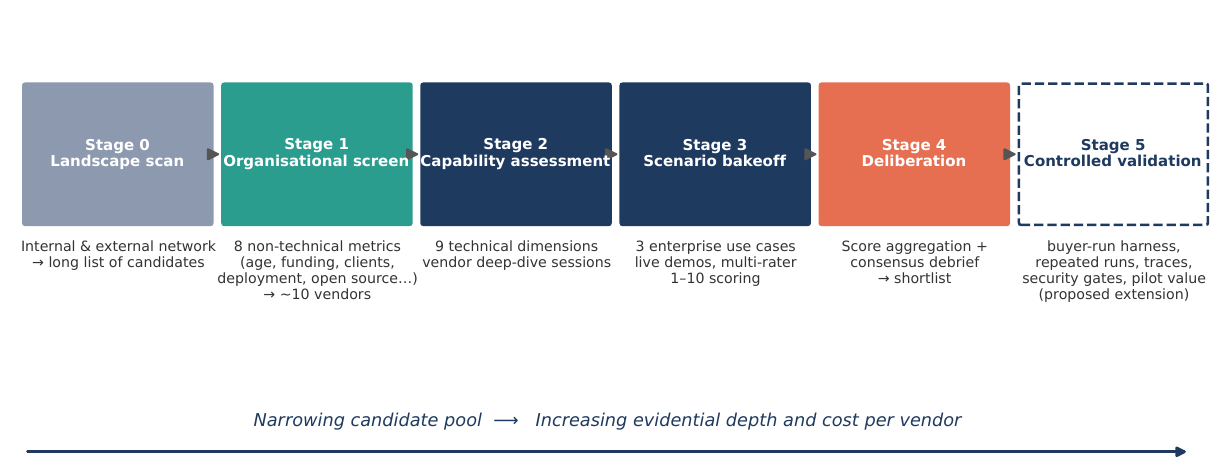}
\caption{Multi-stage evaluation protocol. Each stage narrows the candidate pool while increasing evidential depth and per-vendor cost. Solid: executed; dashed: proposed extension.}\label{fig:protocol}\end{figure*}

Figure~\ref{fig:protocol} summarises the protocol. Solid stages were executed in the bakeoff; the dashed stage is the extension proposed in \S\ref{sec:layers}.

\subsection{Stage 1: organisational screening}
A landscape scan and desk assessment reduced a long list to ten candidates. Each was profiled on eight non-technical attributes-company age, funding stage and amount, key investors, client base, SDKs, deployment options (cloud/on-premise/hybrid) and open-source status-as proxies for vendor viability and integration risk in a multi-year regulated commitment. \status{proposed}

\subsection{Stage 2: the nine-dimension capability taxonomy}
Table~\ref{tab:dims} defines nine dimensions, the guiding question and the architectural component each probes. The mapping lets the rubric be read as more than a checklist: a low reflection score is a statement about the critic loop of Eq.~\eqref{eq:policy}, not merely a missing feature. Two choices merit comment. D7 treats protocol interoperability as first-class, because a bank with dozens of agent initiatives cannot afford bespoke integration per platform. D8 and D9 are where vendor evidence is weakest and the literature most insistent \citep{kapoor2024matter,stroebl2024flaws}; we recommend requiring artefacts (traces, evaluation reports) rather than verbal assurance.

\begin{table*}[t]\centering\small
\caption{Nine-dimension capability taxonomy \status{proposed}, mapped to the reference architecture and the literature each dimension probes.}\label{tab:dims}
\begin{tabular}{@{}L{3.1cm}L{7.6cm}L{2.6cm}L{2.9cm}@{}}\toprule
\textbf{Dimension} & \textbf{Evaluation question} & \textbf{Component} & \textbf{Anchor}\\\midrule
D1 Reflection & Can the agent analyse its outputs, detect errors and revise strategy from execution feedback? & Critic loop & \citep{shinn2023reflexion,huang2024selfcorrect}\\
D2 Tool use & How well does it discover, select, invoke and integrate tools, and recover from tool failures? & Tool layer & \citep{schick2023toolformer,yan2024bfcl}\\
D3 Memory \& retrieval & Does it store, manage and retrieve relevant information within and across interactions? & $m_t$, $\mu$ & \citep{packer2023memgpt,liu2024lost}\\
D4 Planning \& reasoning & Can it decompose goals, reason consistently and adapt plans to changing circumstances? & Policy $\pi$ & \citep{yao2023react,yao2023tot}\\
D5 Multi-agent & Does it support communication, delegation and coordination toward a shared objective? & Orchestration $\Gamma$ & \citep{wu2023autogen,cemri2025fail}\\
D6 Ease of use \& customisation & How accessible is it to developers and business users (docs, API design, learning curve, tailoring)? & Developer experience & \citep{anthropic2024agents}\\
D7 Protocol support & Does it interoperate via MCP, A2A or other open standards? & Interfaces & \citep{anthropic2024mcp,google2025a2a}\\
D8 Quantitative performance & What evidence exists of accuracy, latency and cost on the scenario tasks? & End-to-end & \citep{kapoor2024matter}\\
D9 Evaluation \& explainability & What mechanisms monitor, evaluate and explain behaviour for audit and trust? & Governance plane & \citep{nist2023rmf,mitchell2019modelcards}\\
\bottomrule\end{tabular}\end{table*}

\subsection{Stage 3: scenario design}
Three scenarios span bespoke, regulation-bound analysis to generic, high-volume automation. \textbf{Counterparty credit risk (CCR)}, formulated with business and technical SMEs, combines structured exposure data, unstructured documentation and regulatory logic \citep{bcbs2011basel3}; because vendors are not risk practitioners, the objective was explicitly \emph{adaptability} to a specialised domain, not output correctness (a limitation we return to). \textbf{RAG} \citep{lewis2020rag,gao2023ragsurvey} underpins most enterprise agent workloads and exposes differences in retrieval, grounding and verifiability. \textbf{Process automation} exercises the defining property of agentic AI: goal attainment from minimal instruction within guardrails, with multi-step orchestration and exception handling. CCR stresses D3, D4, D9; RAG stresses D2, D3, D8; automation stresses D1, D2, D4, D5, D7.

\subsection{Scoring model and reliability diagnostics}\label{sec:scoring}
Each rater $r$ in a cross-functional panel $\mathcal R$ (technologists, architects, developers, SMEs, business stakeholders) assigned $x_{v,d,u,r}\in\{1,\dots,10\}$ to vendor $v$ on dimension $d$ for scenario $u$. Scores were aggregated by unweighted means,
\begin{equation}
s_{v,d}=\frac{1}{|\mathcal R_v|}\sum_{r}\frac{1}{|\mathcal U|}\sum_{u}x_{v,d,u,r},\qquad S_v=\sum_{d=1}^{9}w_d\,s_{v,d},\label{eq:score}
\end{equation}
with $w_d=1$ so $S_v\in[9,90]$. Equal weights are defensible for landscape education; adoption decisions should elicit workload-specific weights, e.g.\ by pairwise comparison \citep{saaty1980ahp} (\S\ref{sec:weights}). Because ratings are subjective, the protocol should report inter-rater agreement-ICC(2,$k$) \citep{shrout1979icc} or Krippendorff's $\alpha$ \citep{krippendorff2004}-and internal consistency across dimensions \citep{cronbach1951}. Very high consistency across conceptually distinct dimensions is diagnostic of a \emph{halo effect} \citep{thorndike1920halo}; we test for it in \S\ref{sec:halo}. Per-rater scores were not retained, so inter-rater reliability could not be computed-a required improvement.

\section{Vendor Landscape}\label{sec:landscape}
Ten platforms participated (Table~\ref{tab:landscape}); one invited open-source framework (LangChain/LangGraph) declined and can be assessed independently in future work. A salient axis is the spectrum of developer involvement: pro-code SDKs for control and bespoke integration versus low/no-code builders for ``citizen developers''. Vendors that span the spectrum \emph{with lineage} between modes-prototype in a visual studio, complete in the SDK-preserve both accessibility and auditability of the final artefact, which we regard as the most enterprise-friendly pattern.

\begin{table}[t]\centering\scriptsize
\caption{Evaluated platforms \status{documented}; vendor material current at evaluation. Microsoft was scored but not profiled in the original programme; its entry reflects public documentation.}\label{tab:landscape}
\begin{tabular}{@{}L{1.8cm}L{1.7cm}L{1.7cm}L{1.8cm}@{}}\toprule
\textbf{Platform} & \textbf{Licensing} & \textbf{Modality} & \textbf{Heritage}\\\midrule
H2O.ai (h2oGPTe) & Proprietary + OSS components & Low-code, generated Python; SDK & Predictive ML $\to$ GenAI; sovereign/air-gapped\\
Google (Vertex AI Agent Builder, ADK) & Managed + OSS ADK & No-code console; pro-code ADK & Hyperscaler\\
Microsoft (Foundry Agent Service, Semantic Kernel, AutoGen) & Managed + OSS SDKs & Pro- and low-code & Hyperscaler\\
Writer & Proprietary (Palmyra) & Visual builder; SDK & Enterprise generative writing\\
Letta & OSS core + platform & Pro-code, API-first & Research (MemGPT)\\
Glean & Proprietary & NL builder & Enterprise search\\
CrewAI & OSS + Enterprise Suite & Python; Studio & Multi-agent framework\\
UiPath & Proprietary & Low-code (Agent Builder, Maestro) & RPA\\
Portia AI & OSS SDK + platform & Pro-code & Plan-first framework\\
Squid AI & Proprietary BaaS & Low-code Studio; SDKs & Backend-as-a-service\\
\bottomrule\end{tabular}\end{table}

\section{Results}\label{sec:results}
\subsection{Aggregate scores and a three-tier structure}
\begin{figure}[t]\centering\includegraphics[width=\columnwidth]{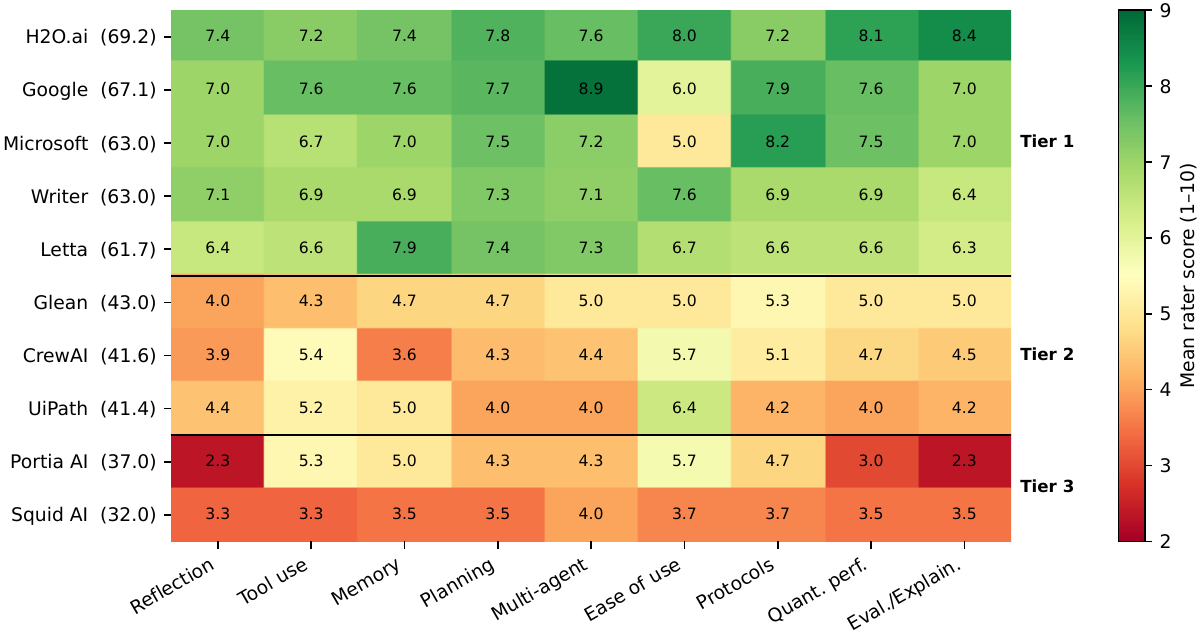}
\caption{Heat map of mean panel scores; horizontal rules mark the three tiers.}\label{fig:heat}\end{figure}

\begin{figure*}[t]\centering\includegraphics[width=0.95\textwidth]{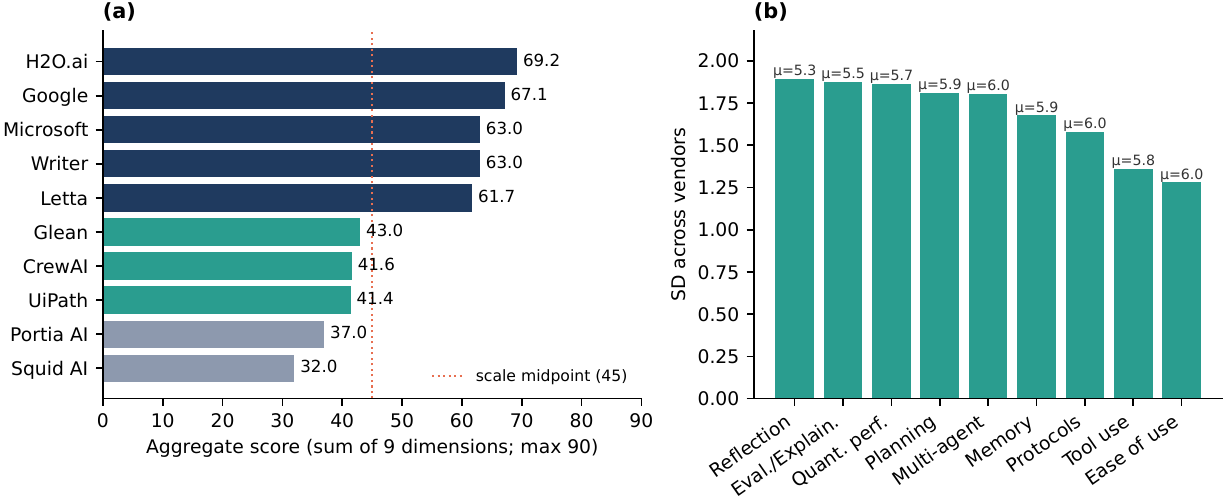}
\caption{(a) Aggregate scores with tier colouring. (b) Cross-vendor standard deviation per dimension (discriminative power); $\mu$ is the cross-vendor mean.}\label{fig:totals}\end{figure*}

Table~\ref{tab:scores} and Fig.~\ref{fig:heat} report mean panel scores. They derive exclusively from live demonstrations and accompanying presentations under assisted conditions (\S\ref{sec:threats}); they reflect \status{observed} performance in the Lab's context, not market reputation. Totals cluster into three tiers separated by gaps far larger than within-tier differences (Fig.~\ref{fig:totals}a): \textbf{Tier~1} (61.7--69.2) H2O.ai, Google, Microsoft, Writer, Letta; \textbf{Tier~2} (41.4--43.0) Glean, CrewAI, UiPath; \textbf{Tier~3} (32.0--37.0) Portia AI, Squid AI. The 18.7-point gap between Tiers~1 and~2 dwarfs the 7.5-point spread within Tier~1. The numerical ranking coincided with the consensus reached in panel deliberation-reassuring, but not independent, since the same people produced both.

\begin{table*}[t]\centering\small
\caption{Mean panel scores (1--10) by platform and dimension, ordered by total (max 90). Dimension codes as in Table~\ref{tab:dims}.}\label{tab:scores}
\begin{tabular}{@{}lrrrrrrrrrr@{}}\toprule
\textbf{Platform} & D1 & D2 & D3 & D4 & D5 & D6 & D7 & D8 & D9 & \textbf{Total}\\\midrule
H2O.ai & 7.44 & 7.22 & 7.44 & 7.78 & 7.56 & 8.00 & 7.22 & 8.11 & \textbf{8.44} & \textbf{69.2}\\
Google & 7.00 & 7.57 & 7.57 & 7.71 & \textbf{8.86} & 6.00 & 7.86 & 7.57 & 7.00 & 67.1\\
Microsoft & 7.00 & 6.67 & 7.00 & 7.50 & 7.17 & 5.00 & \textbf{8.17} & 7.50 & 7.00 & 63.0\\
Writer & 7.14 & 6.86 & 6.86 & 7.29 & 7.14 & 7.57 & 6.86 & 6.86 & 6.43 & 63.0\\
Letta & 6.43 & 6.57 & \textbf{7.86} & 7.43 & 7.29 & 6.71 & 6.57 & 6.57 & 6.29 & 61.7\\\midrule
Glean & 4.00 & 4.33 & 4.67 & 4.67 & 5.00 & 5.00 & 5.33 & 5.00 & 5.00 & 43.0\\
CrewAI & 3.90 & 5.40 & 3.60 & 4.30 & 4.40 & 5.70 & 5.10 & 4.70 & 4.50 & 41.6\\
UiPath & 4.40 & 5.20 & 5.00 & 4.00 & 4.00 & 6.40 & 4.20 & 4.00 & 4.20 & 41.4\\\midrule
Portia AI & 2.33 & 5.33 & 5.00 & 4.33 & 4.33 & 5.67 & 4.67 & 3.00 & 2.33 & 37.0\\
Squid AI & 3.33 & 3.33 & 3.50 & 3.50 & 4.00 & 3.67 & 3.67 & 3.50 & 3.50 & 32.0\\\midrule
Mean & 5.30 & 5.85 & 5.85 & 5.85 & 5.98 & 5.97 & 5.97 & 5.68 & 5.47 & \\
SD across vendors & 1.87 & 1.41 & 1.68 & 1.75 & 1.78 & 1.28 & 1.62 & 1.85 & 1.94 & \\
$\rho$ with total & 0.95 & 0.93 & 0.95 & 0.99 & 0.96 & 0.66 & 0.94 & 0.98 & 0.96 & \\
\bottomrule\end{tabular}\end{table*}

\subsection{Dimension-level patterns}
Four patterns are visible (Table~\ref{tab:scores}, Fig.~\ref{fig:totals}b; radar profiles in Appendix~A (supplementary PDF)).
\textbf{Multi-agent (D5) and protocol support (D7) favour hyperscalers.} Google recorded the highest cell in the matrix (8.86 on D5) and Microsoft the highest D7 (8.17), consistent with both firms' stewardship of A2A and MCP investment.
\textbf{Memory (D3) favours the research-derived platform.} Letta, built on MemGPT, achieved the top memory score despite a mid-Tier-1 total, showing that the taxonomy surfaces specialist strength that aggregates obscure.
\textbf{Ease of use (D6) is the least discriminating and most independent dimension} (SD 1.28; $\rho=0.66$ with total versus $\geq0.93$ for every other dimension). Microsoft's D6 (5.00) was the lowest in Tier~1 and UiPath's (6.40) the highest in Tier~2-evidence that raters could and did decouple usability from capability.
\textbf{Reflection (D1) and evaluation/explainability (D9) have the lowest means and widest ranges.} These are the dimensions on which the literature finds systems weakest \citep{huang2024selfcorrect,kapoor2024matter} and regulators are most demanding; they should carry heavier weights in adoption decisions.

\subsection{Internal consistency and the halo hypothesis}\label{sec:halo}
Treating the nine dimensions as items and ten platforms as cases, Cronbach's $\alpha=0.98$ and eight of nine dimensions correlate with the total at $\rho\geq0.93$. Genuinely distinct capabilities-memory management and protocol support, say-should not co-vary this tightly across products of such different heritage. The parsimonious explanation is a strong common factor driven by overall demonstration impression: a halo effect \citep{thorndike1920halo}. This does not invalidate the tiering, which deliberation corroborated, but it implies the rubric as administered measured roughly one-and-a-half constructs rather than nine. Two protocol changes follow: require dimension-specific evidence artefacts, and score dimensions in separate sessions or by specialist sub-panels to break the shared impression.

\subsection{Weighting sensitivity}\label{sec:weights}
Totals were recomputed under a \emph{governance-first} profile (D9$\times$2, D7$\times$1.5, D6$\times$0.5) and a \emph{developer-first} profile (D6$\times$2, D2$\times$1.5, D9$\times$0.5), each normalised to sum to nine. Tier membership was unchanged under both. Within-tier order shifted modestly (developer-first: Writer 63.7 overtakes Microsoft 61.1; UiPath 43.5 overtakes CrewAI and Glean). Tiers, not ranks, are the robust finding.

\subsection{Scenario-level disaggregation}
For the top three platforms (supplementary Appendix~B, Table~A1), H2O.ai's profile is flat across scenarios with a peak on D9 in CCR (8.75); Google's D5 is uniformly high (8.67--9.00) while its D6 is uniformly capped at 6.00-a systematic rather than scenario-specific concern; Microsoft's scores are almost invariant across scenarios, which is itself informative: raters saw little scenario-specific adaptation.

\section{Threats to Validity}\label{sec:threats}
Table~\ref{tab:threats} states the threats precisely rather than claiming more than the evidence supports. Two deserve emphasis. \emph{Assisted conditions}: the Lab helped vendors prepare, so the object measured is demonstrated capability under assisted conditions, not out-of-the-box performance. \emph{Unequal panels}: score granularity implies panels of roughly 3--10 raters per vendor (e.g.\ $7.44=67/9$, $4.33=13/3$; an inference from decimal patterns), inflating variance for small panels.

\begin{table*}[t]\centering\small
\caption{Threats to validity and mitigations.}\label{tab:threats}
\begin{tabular}{@{}L{2.7cm}L{6.2cm}L{3.0cm}L{4.6cm}@{}}\toprule
\textbf{Threat} & \textbf{Description} & \textbf{Applied} & \textbf{Recommended}\\\midrule
Presentation confound & Demonstration clarity influenced perceived capability & Preparatory sessions; set-up support & Blind, hands-on evaluation on buyer infrastructure; artefact-based scoring\\
Halo effect & $\alpha=0.98$ across dimensions (\S\ref{sec:halo}) & None & Per-dimension evidence; sub-panels; randomised order\\
Unequal panel size & $\sim$3--10 raters per vendor (inferred) & None & Fixed panel; report $n$ and CIs\\
Ordinal as interval & 1--10 ratings summed as if equally spaced & Anchors (supplementary Appendix~E) & Medians and IRR; anchored scales\\
No ground truth (CCR) & Adaptability, not correctness, assessed & Scope stated & SME gold answers; end-state checks\\
Vendor self-selection & One invited framework declined & Noted & Evaluate OSS frameworks independently\\
Single organisation & One institution, risk appetite, estate & Noted & Replicate across institutions\\
Construct validity & ``Agentic capability'' is workload-dependent & Weighting sensitivity & Gate-then-score with per-workload weights\\
Temporal validity & Platforms evolve monthly & Dates recorded & Version pinning; periodic re-run\\
Cost/latency & D8 relied on vendor-reported figures & None & Cost-of-pass accounting \citep{kapoor2024matter}\\
Safety & No adversarial testing & None & Indirect-injection red-teaming \citep{greshake2023injection,debenedetti2024agentdojo}\\
\bottomrule\end{tabular}\end{table*}

\section{A Gate-then-Score Layered Model}\label{sec:layers}
\begin{figure}[t]\centering\includegraphics[width=\columnwidth]{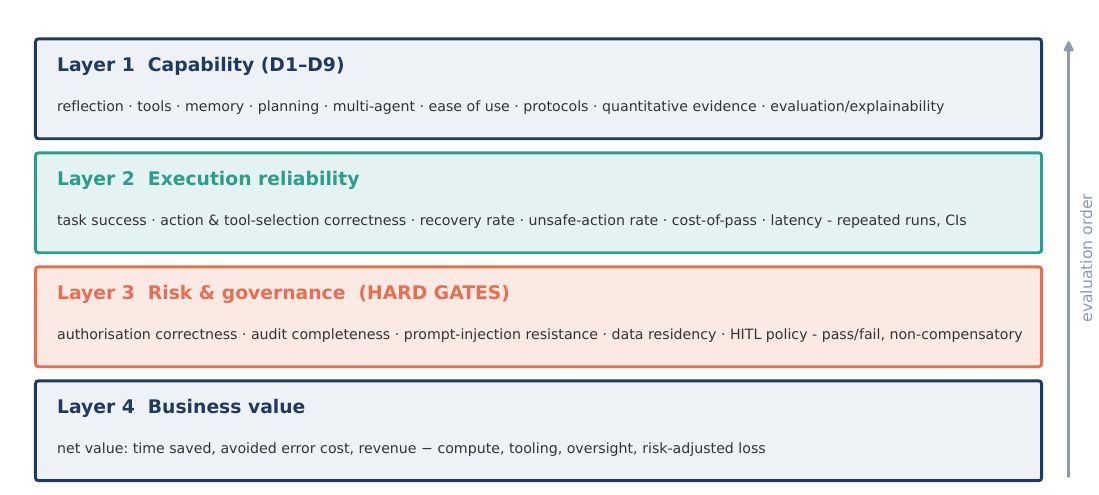}
\caption{Layered evaluation model. Gates are pass/fail and precede scoring; business value is measured in pilot, not demonstration.}\label{fig:layers}\end{figure}

The nine dimensions describe \emph{capability}. Deployment additionally requires evidence about \emph{execution reliability}, \emph{risk and governance} and \emph{business value}, and these are not interchangeable: no capability score should compensate for a failed authorisation control. We propose the layered model of Fig.~\ref{fig:layers}: Layer~3 is evaluated first as non-compensatory gates, Layers~1--2 are scored within the admissible set, Layer~4 is validated in pilot. \status{proposed}

\paragraph{Decision rule.}
Let $\mathcal G=\{G_1,\dots,G_K\}$ be binary gates (authorisation correctness, audit-trace completeness, injection resistance, data residency, human-approval policy, minimum reliability), $x_{v,d}=(s_{v,d}-1)/9\in[0,1]$ normalised scores, and $w_\rho$ a weight vector for risk tier $\rho$ with $\sum_d w_{\rho,d}=1$. Then
\begin{equation}\label{eq:eas}
\begin{aligned}
\mathrm{EAS}(v,\rho)&=\textstyle\sum_d w_{\rho,d}\,x_{v,d},\\
v^\star&=\operatorname*{arg\,max}_{v:\,G_k(v)=1\ \forall k}\mathrm{EAS}(v,\rho)\\
&\quad\text{s.t.}\ C(v)\le C_{\max},\ L(v)\le L_{\max},
\end{aligned}
\end{equation}
with $C,L$ cost and latency constraints. Eq.~\eqref{eq:eas} replaces an unconstrained ranking with a constrained optimisation; a platform failing any gate is inadmissible regardless of EAS. Applied retrospectively, the bakeoff evaluated Layer~1 only.

\paragraph{Trace-level reliability (Layer 2).}
Trajectories are stochastic, so Layer-2 metrics must be estimated from $K$ repeated runs with uncertainty. For task $\tau$ with outcomes $y_k\in\{0,1\}$ and per-run cost $c_k$,
\begin{equation}\label{eq:l2}
\hat p_\tau=\tfrac1K\textstyle\sum_k y_k,\quad \mathrm{pass}^k\approx\hat p_\tau^{\,k},\quad \mathrm{CoP}_\tau=\frac{\sum_k c_k}{\sum_k y_k},
\end{equation}
with a Wilson 95\% interval on $\hat p_\tau$ \citep{wilson1927}, preferred for small $K$; pass$^k$ measures consistency rather than best-case ability \citep{yao2024taubench}, and cost-of-pass prices a success inclusive of failed attempts \citep{kapoor2024matter}. Table~\ref{tab:metrics} defines the metric set; every metric is computed from the execution trace, so trace retention is a precondition. Tools must be tagged \emph{before} evaluation with permission scope, statefulness, reversibility and side-effect severity-a read-only search and a payment API are not equivalent actions \citep{shavit2023governing,nist2024genai}.

\begin{table}[t]\centering\scriptsize
\caption{Trace-level metrics for execution reliability and governance gates \status{proposed}.}\label{tab:metrics}
\begin{tabular}{@{}L{2.2cm}L{3.6cm}L{1.6cm}@{}}\toprule
\textbf{Metric} & \textbf{Definition} & \textbf{Layer}\\\midrule
Task success $\hat p$ & Runs meeting the pre-registered success predicate on final state & L2\\
Consistency pass$^k$ & Probability that $k$ independent runs all succeed & L2\\
Action correctness & State-changing actions that were necessary and correctly parameterised & L2\\
Tool-selection accuracy & Invocations choosing the appropriate tool for the sub-goal & L2\\
Recovery rate & Injected/natural tool failures after which the run still succeeds & L2\\
Unsafe-action rate & Out-of-scope state changes per 1,000 actions, vs.\ permission tags & L3 gate (0 for irreversible)\\
Injection susceptibility & Adversarial documents/tool outputs that alter behaviour \citep{debenedetti2024agentdojo} & L3 gate\\
Cost-of-pass; latency p50/p95 & Cost per success; wall-clock distribution & L2 / constraint\\
Human-intervention rate & Escalations or approvals per task & L2 $\to$ L4\\
\bottomrule\end{tabular}\end{table}

\begin{algorithm}[t]\caption{Gate-then-score enterprise evaluation protocol}\label{alg:protocol}
\begin{algorithmic}[1]
\REQUIRE platforms $V$; scenarios $U$ with success predicates; gates $\mathcal G$; tier $\rho$; weights $w_\rho$; runs $K$
\ENSURE admissible set $A$ ranked by EAS with Layer-2 CIs
\FOR[Stages 1--2]{$v\in V$}
  \STATE record organisational attributes; collect per-dimension evidence artefacts
  \STATE $s[v,d]\gets$ panel median of anchored ratings for each $d$; report IRR
\ENDFOR
\FOR[Stages 3, 5: controlled harness]{$v\in V,\ u\in U$}
  \STATE pin model, prompts, tool schemas, data; tag tools (scope, reversibility, severity)
  \FOR{$k=1..K$} \STATE run $v$ on $u$; store trace $t[v,u,k]$; $y_k\gets\mathrm{predicate}(\text{final state})$ \ENDFOR
  \STATE compute $\hat p$, pass$^k$, CoP, recovery, unsafe-action rate from traces \COMMENT{Eq.~\eqref{eq:l2}}
  \STATE run adversarial variants of $u$ (indirect injection, authorisation probes)
\ENDFOR
\STATE $A\gets\{v: G_k(v)=1\ \forall k\}$ \COMMENT{non-compensatory gates}
\FOR{$v\in A$} \STATE $\mathrm{EAS}[v]\gets\sum_d w_{\rho,d}(s[v,d]-1)/9$; attach Layer-2 CIs \ENDFOR
\STATE shortlist $\gets$ top of $A$ s.t.\ $C(v)\le C_{\max}$, $L(v)\le L_{\max}$
\STATE pilot shortlist; measure Layer-4 terms (Eq.~\eqref{eq:nv}); repeat lines 5--10 on version change
\end{algorithmic}\end{algorithm}

Algorithm~\ref{alg:protocol} states the protocol. Lines 1-4 correspond to what the bakeoff executed; lines 5--16 are the proposed extension. The asymmetry is deliberate: gating before scoring means a platform can be excluded on a single control failure, which is correct for state-changing systems in a regulated setting.

\section{Discussion}\label{sec:discussion}

\subsection{Connecting capability to business value}
Layer~4 asks whether a workflow $j$ is worth deploying at all. We propose the accounting identity
\begin{equation}
\mathrm{NV}_j=T_j+E_j+R_j-C^{\text{compute}}_j-C^{\text{tools}}_j-O_j-\mathbb E[\ell_j],\label{eq:nv}
\end{equation}
where $T$ is labour time saved at loaded cost, $E$ avoided error cost, $R$ incremental revenue, $C$ compute and tooling cost, $O$ oversight cost (approvals, exceptions, audit) and $\mathbb E[\ell]$ the risk-adjusted expected loss from erroneous state changes. In financial services the last two terms often dominate: an automation that saves analyst time but raises the probability of a material misstatement has negative net value. All terms are measured from pilot data; none can be read from a demonstration. \status{hypothesis}

\subsection{Lessons for practitioners}
Three changes raise the evidential standard of enterprise bakeoffs without prohibitive cost: (1)~move from vendor-run to buyer-run demonstrations on controlled infrastructure with pinned task definitions; (2)~require dimension-specific artefacts-trace logs, evaluation dashboards, memory dumps-so raters score evidence, not narration; (3)~retain per-rater data and report reliability statistics alongside means. With these, the taxonomy becomes a repeatable instrument rather than a structured opinion survey.


\section{Conclusion}
Enterprises are adopting a class of technology whose defining property-autonomous, state-changing action-is what regulated institutions must control most tightly. We formalised a bakeoff into a reusable protocol: an organisational screen, a nine-dimension taxonomy grounded in the literature, a scenario portfolio from bespoke analysis to autonomous execution, and a scoring model with reliability diagnostics. Applied to ten platforms it produced a robust three-tier structure in which hyperscaler ecosystems and research-derived platforms led, while exposing-through $\alpha=0.98$-the halo risk inherent in demonstration-based scoring. The specialist strengths it surfaced (memory at Letta, protocol depth at Microsoft, evaluation tooling at H2O.ai, action grounding at UiPath) are a reminder that ``best platform'' is workload-relative. We then extended the rubric into a gate-then-score layered model with trace-level metrics and a reproducibility protocol so the next iteration can measure rather than rate. The most consequential finding is procedural: current enterprise evaluation practice measures presentation more than capability, and the remedy-buyer-controlled tasks, evidence artefacts and measured reliability-is within reach.

\bibliography{main}
\bibliographystyle{mlsys2025}

\appendix 

\appendix
\section*{Supplementary Appendix}
\label{app:appendix}


\subsection*{A. Radar Profiles of Tier-1 Platforms}\label{app:radar}
\begin{figure}[H]\centering\includegraphics[width=0.92\columnwidth]{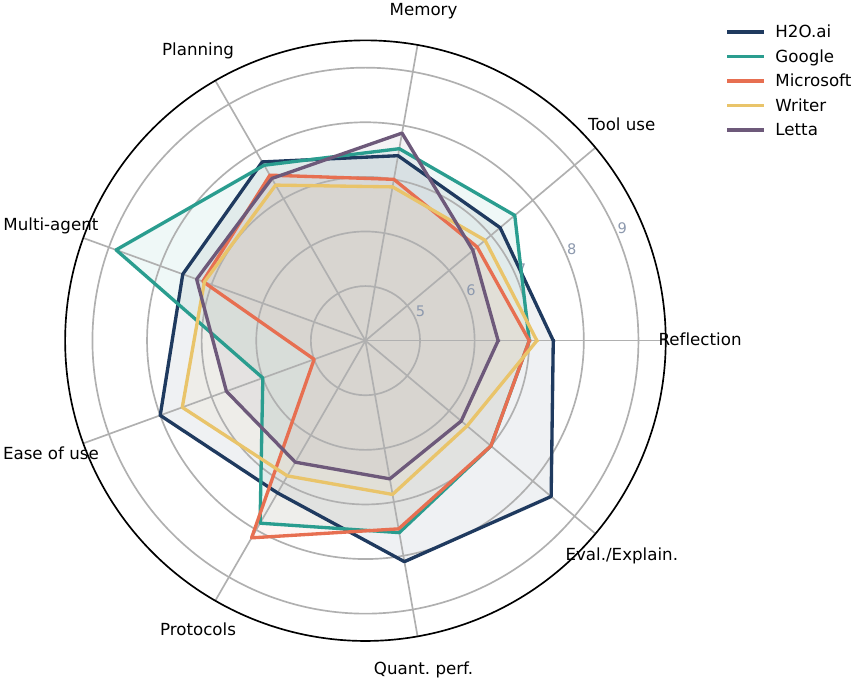}
\caption{Capability profiles of the five Tier-1 platforms (axis range 4--9.5 to emphasise differences).}\end{figure}

\subsection*{B. Platform Profiles}\label{app:profiles}
Each profile separates what is \status{documented} in vendor documentation from what was \status{observed} in the bakeoff demonstration, and closes with the panel scores. Vendor-reported figures that we could not verify are marked. Platforms evolve monthly; where documentation now differs from what was demonstrated, we say so.

\paragraph{B.1 H2O.ai (h2oGPTe).}
\status{documented} Multi-agent platform (general, data-science and vertical agents) that calls H2O's predictive models (Driverless AI) from generative workflows; deploys air-gapped, on-premise, VPC or managed cloud; guardrails and PII controls; third-party LLM support \citep{h2o2024agentic,h2o2026docs}. Vendor-reported 79.7\% on GAIA (June 2025; not independently verified). \status{observed} Low-code builder emitting editable Python; web research, database access, code execution, self-critique; citation-based RAG verification. Highest total (69.2) and highest D9 (8.44).

\paragraph{B.2 Google (Vertex AI Agent Builder).}
\status{documented} Open-source, model-agnostic Agent Development Kit (ADK) for multi-agent systems; Agent Designer (no-code) with hand-off to ADK; managed Agent Engine runtime with Sessions, Memory Bank, Code Execution, evaluation and tracing; also hosts LangChain, LangGraph, AG2 and LlamaIndex agents; A2A template (preview); RAG Engine and Vertex AI Search grounding; agent identity via IAM and Agent Engine threat detection \citep{google2026agentbuilder,google2026adk}. \status{observed} Strongest multi-agent demonstration in the field. Highest D5 (8.86); D6 capped at 6.00 in every scenario, reflecting the breadth of the cloud surface.

\paragraph{B.3 Microsoft (Foundry Agent Service, Agent Framework).}
\status{documented} Foundry Agent Service: managed runtime with Python/C\#/TypeScript SDKs, MCP tool integration, built-in tools (Fabric, SharePoint, Azure AI Search), vector store and conversation state. Microsoft Agent Framework (open source) is the declared successor to Semantic Kernel and AutoGen: session state, middleware, telemetry, graph workflows (sequential, concurrent, hand-off, group chat, Magentic), and MCP, A2A and OpenAPI interoperability \citep{ms2026agentframework,ms2026foundry}. \status{observed} Scored but not profiled in the original programme; the toolset demonstrated was fragmented (SK + AutoGen), which the unified framework has since addressed. Highest D7 (8.17); lowest Tier-1 D6 (5.00).

\paragraph{B.4 Writer (AI Studio).}
\status{documented} Full-stack platform: no-code Agent Builder, Writer Framework (code) and API; proprietary Palmyra models (X5, Fin, Med, Vision) plus external models via Bedrock; Knowledge Graph RAG with inline citations and connectors; guardrails (content safety, PII, compliance); platform and agent observability; MCP server; LangChain and AWS Strands integrations \citep{writer2026docs}. \status{observed} Four-level capability model (assistive, knowledge, action, multi-agent); ``Action AI'' agents executing in sandboxed Linux environments (vendor description). Balanced profile; strong D6 (7.57).

\paragraph{B.5 Letta.}
\status{documented} Open-source server and cloud from the MemGPT authors \citep{packer2023memgpt}: self-editing memory split into in-context memory blocks and out-of-context archival memory; blocks shareable across agents; sleep-time agents that rewrite memory in the background; Agent Development Environment (ADE) exposing memory, context window and tool calls; MCP tool servers; Agent File (.af) for portable agent state; Letta Evals; REST/Python/TypeScript SDKs; self-host or cloud \citep{letta2026docs}. \status{observed} Agents migrating between model providers without losing state. Top D3 (7.86); weaker D1 (6.43) and D9 (6.29).

\paragraph{B.6 Glean.}
\status{documented} Agent layer over permission-aware enterprise search: Agent Builder with a Plan + Execute step, event and schedule triggers, model-agnostic; Glean acts as both MCP host (admin-gated remote servers) and managed MCP server exposing search, chat and agents as tools; Glean Protect enforces source permissions on every action; LangChain, LangGraph and OpenAI Agents SDK integrations \citep{glean2026mcp,glean2025open}. \status{observed} Uniform mid-range scores (4.00--5.33); its strength, permission-aware knowledge access, sits largely outside the rubric's emphasis on autonomous execution.

\paragraph{B.7 CrewAI.}
\status{documented} Open-source Python framework: Agents (role, goal, tools), Tasks, Crews with sequential or hierarchical process (manager agent delegates and validates); Flows for event-driven, stateful, resumable workflows with human-in-the-loop; unified Memory with hierarchical scopes; knowledge, guardrails and observability. Enterprise (AMP) adds a console, triggers (Gmail, Slack, Salesforce, HubSpot), RBAC and environment management \citep{crewai2026docs}. \status{observed} Demonstrated memory and reflection depth was limited; the unified memory system post-dates the bakeoff. Tier 2 (41.6) despite strong community standing: a clear instance of the presentation confound.

\paragraph{B.8 UiPath.}
\status{documented} Agent Builder: low-code canvas in Studio with Autopilot natural-language authoring, context grounding, evaluations and an Agent Score; Maestro orchestrates agents, RPA robots and people in one process, with agent decisions routing execution; tools include connectors, RPA/API workflows, document extraction and MCP servers; AI Trust Layer provides model governance, audit logs, RBAC and PII filtering; agents, Maestro and MCP servers also available on-premise in Automation Suite \citep{uipath2026agentbuilder,uipath2026maestro}. \status{observed} Distinctive grounding in action against legacy systems without APIs. Strongest Tier-2 D6 (6.40); weakest D4 and D5 (4.00), consistent with a workflow-centric heritage.

\paragraph{B.9 Portia AI.}
\status{documented} Open-source Python SDK for ``predictable, stateful, authenticated'' agents: a planner emits an explicit Plan; a PlanRun carries all execution state; Clarifications pause a run for human input, approval or authentication; execution hooks before/after each step and tool call; tool registries from MCP servers, local tools or Portia cloud tools; Portia Cloud dashboard logs runs; Steel Thread provides offline and online evals (latency, LLM-judge, tool-call correctness) \citep{portia2026hooks,portia2026steelthread}. \status{observed} Structured agent-to-human hand-overs. Lowest D1 and D9 (2.33) despite an auditability thesis; most plausibly demonstration scope, since the documented hooks and evals target exactly these dimensions.

\paragraph{B.10 Squid AI.}
\status{documented} Serverless backend-as-a-service with built-in vector store, embeddings and ``universal RAG'' over structured and unstructured data; connects to any database or API in cloud or on-premise; knowledge, system and data agents that inherit source permissions; client and backend SDKs and AI functions for action-taking \citep{squid2026}. \status{observed} Lead agent orchestrating expert agents; low-code studio; RAG over tables and images. Lowest scores on every dimension (32.0 total).

\subsection*{C. Scenario-Level Scores}\label{app:scenario}
\begin{figure}[H]\centering\includegraphics[width=\columnwidth]{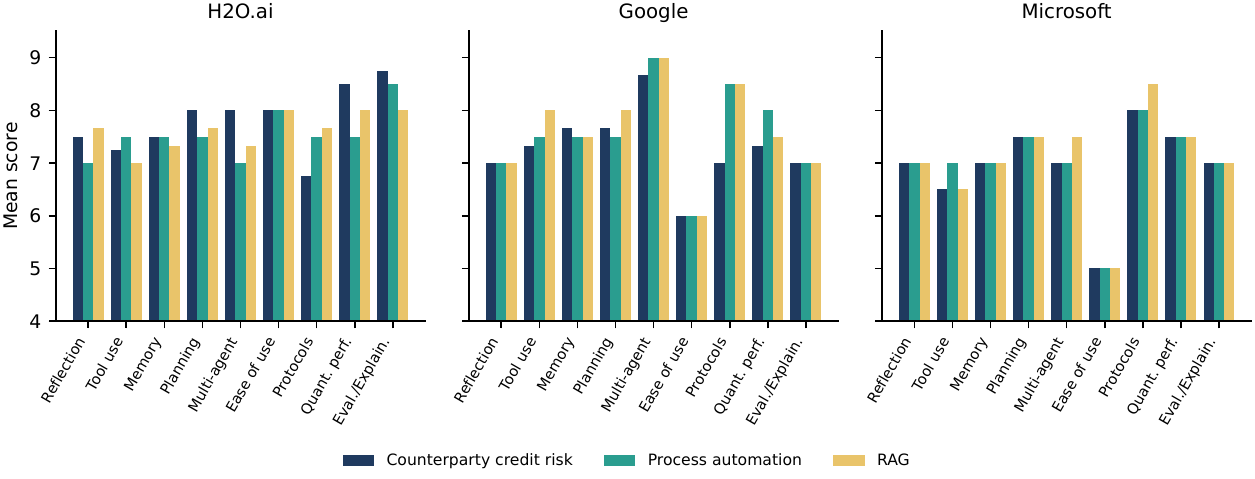}
\caption{Scenario-level scores for the three highest-scoring platforms.}\label{fig:scenario}\end{figure}

Table~\ref{tab:scenario} and Figure~\ref{fig:scenario} disaggregate the three highest-scoring platforms by scenario. H2O.ai is flat across scenarios with a peak on D9 in CCR; Google's D5 is uniformly high while its D6 is capped at 6.00 in every scenario; Microsoft's profile is almost invariant, indicating little scenario-specific adaptation in demonstration.

\setcounter{table}{2}
\begin{table*}[t]\centering\scriptsize\setlength{\tabcolsep}{5pt}
\caption{Behavioural anchors for the nine dimensions (intermediate scores interpolate).}\label{tab:anchors}
\begin{tabular}{@{}L{2.1cm}L{2.5cm}L{2.9cm}L{3.5cm}L{4.0cm}@{}}\toprule
\textbf{Dimension} & \textbf{2 (absent)} & \textbf{5 (basic)} & \textbf{8 (strong)} & \textbf{10 (state of the art)}\\\midrule
D1 Reflection & No self-evaluation & Retries on error & Verbal critique revises plan with improved outcome & External-verifier-driven repair loops with measured gains\\
D2 Tool use & Hard-coded calls & Static list; no recovery & Dynamic selection; typed schemas; graceful failure & Tool discovery; MCP-native; sandboxed execution with audit\\
D3 Memory & Context only & Session store & Episodic + semantic stores; observable & Self-editing long-term memory; model migration\\
D4 Planning & Single-shot & Linear decomposition & Hierarchical plans; replanning & Inspectable plan objects; conditional branches; verification\\
D5 Multi-agent & None & Sequential hand-off & Hierarchical delegation with validation & Roles, negotiation, A2A; failure containment\\
D6 Ease of use & Undocumented & SDK only & Studio + SDK with lineage; good docs & Business-user authoring with governed promotion\\
D7 Protocols & Proprietary & Partial MCP & MCP client and server; A2A roadmap & MCP + A2A native; open schemas\\
D8 Quant.\ performance & Anecdotal & Vendor benchmarks & Reproducible metrics on buyer tasks & Cost-of-pass Pareto data; latency SLOs\\
D9 Eval.\ \& explainability & Black box & Logs & Trace viewer; eval harness; per-step explanations & Continuous evaluation; policy audit; regulator-ready reports\\
\bottomrule\end{tabular}\end{table*}
\setcounter{table}{0}

\begin{table}[H]\centering\scriptsize\setlength{\tabcolsep}{4.5pt}
\caption{Scenario-level mean scores for the three highest-scoring platforms (CCR: counterparty credit risk; PA: process automation).}\label{tab:scenario}
\begin{tabular}{@{}llrrrrrrrrr@{}}\toprule
\textbf{Platform} & \textbf{Scen.} & D1 & D2 & D3 & D4 & D5 & D6 & D7 & D8 & D9\\\midrule
\multirow{3}{*}{H2O.ai} & CCR & 7.50 & 7.25 & 7.50 & 8.00 & 8.00 & 8.00 & 6.75 & 8.50 & 8.75\\
 & PA & 7.00 & 7.50 & 7.50 & 7.50 & 7.00 & 8.00 & 7.50 & 7.50 & 8.50\\
 & RAG & 7.67 & 7.00 & 7.33 & 7.67 & 7.33 & 8.00 & 7.67 & 8.00 & 8.00\\\midrule
\multirow{3}{*}{Google} & CCR & 7.00 & 7.33 & 7.67 & 7.67 & 8.67 & 6.00 & 7.00 & 7.33 & 7.00\\
 & PA & 7.00 & 7.50 & 7.50 & 7.50 & 9.00 & 6.00 & 8.50 & 8.00 & 7.00\\
 & RAG & 7.00 & 8.00 & 7.50 & 8.00 & 9.00 & 6.00 & 8.50 & 7.50 & 7.00\\\midrule
\multirow{3}{*}{Microsoft} & CCR & 7.00 & 6.50 & 7.00 & 7.50 & 7.00 & 5.00 & 8.00 & 7.50 & 7.00\\
 & PA & 7.00 & 7.00 & 7.00 & 7.50 & 7.00 & 5.00 & 8.00 & 7.50 & 7.00\\
 & RAG & 7.00 & 6.50 & 7.00 & 7.50 & 7.50 & 5.00 & 8.50 & 7.50 & 7.00\\
\bottomrule\end{tabular}\end{table}

\begin{table}[H]\centering\scriptsize\setlength{\tabcolsep}{3pt}
\caption{Supervised ML extraction versus agentic document extraction.}\label{tab:ade}
\begin{tabular}{@{}L{1.55cm}L{1.95cm}L{2.05cm}L{1.95cm}@{}}\toprule
\textbf{Dimension} & \textbf{Supervised ML} & \textbf{GenAI ADE} & \textbf{Status}\\\midrule
Input modality & Structured/semi-structured layouts & Text, tables, images, transcripts jointly & Established \citep{mathew2021docvqa}\\
New formats & Retraining with labels & Zero-/few-shot & Model-level established; production robustness is a hypothesis\\
Accuracy profile & High, in-distribution & Broad; varies with layout complexity & Observed; not benchmarked\\
Context awareness & Field capture & Roles, dependencies, intent & Observed\\
Human oversight & Post-hoc QA & Confidence-gated checkpoints & Proposed\\
\bottomrule\end{tabular}\end{table}

\paragraph{C.1 Vendor observation (Landing AI).} Vision APIs extract structured data from PDFs and images without templates or training, returning hierarchical JSON with page and coordinate references; vendor-reported median processing time reduction from 135\,s to 8\,s \status{documented, not verified}. A qualitative comparison with direct LLM extraction (Gemini) suggested the LLM excels at contextual reasoning while a specialised extractor embedded in an agentic pipeline offers more control over cost, checkpoints and vendor independence-a design \status{hypothesis}, not a measured result.

\paragraph{C.2 Publication-grade ADE metrics.} A labelled corpus stratified by type, layout complexity and modality; field-level precision/recall/F1; table- and layout-level accuracy; bounding-box or citation correctness; abstention quality (precision of the review flag); human-review rate; latency and cost per page; robustness under distribution shift; all with confidence intervals.

\subsection*{D. Proposed Scoring Anchors}\label{app:anchors}
The original programme used an unanchored 1--10 scale. To reduce rater drift and make scores comparable across panels, Table~\ref{tab:anchors} proposes behavioural anchors at four points on each dimension; intermediate scores interpolate. Anchors are illustrative and should be refined per organisation and workload before use.

\end{document}